\documentclass{article}

\usepackage[preprint]{neurips_2022}
\usepackage{graphicx}
\usepackage{pgfplots}
\usepackage{algorithm}
\usepackage{algpseudocode}

\usepackage[utf8]{inputenc} 
\usepackage[T1]{fontenc}    
\usepackage{url}            
\usepackage{booktabs}       
\usepackage{amsfonts}       
\usepackage{nicefrac}       
\usepackage{microtype}      
\usepackage{xcolor}         

\title{Uncertainty-Aware Perceiver}

\author{%
  EuiYul Song\thanks{Recent Advances in Deep Learning, Seoul, South Korea, 2022. Copyright 2022 by the author(s).} \\
  Department of Artificial Intelligence \\
  KAIST \\
  Daejeon, Korea \\
  \texttt{thddmlduf@kaist.ac.kr} \\
}

\begin{document}

\maketitle

\begin{abstract}
 The Perceiver makes few architectural assumptions about the relationship among its inputs with quadratic scalability on its memory and computation time. Indeed, the Perceiver model outpaces or is competitive with ResNet-50 and ViT in terms of accuracy to some degree. However, the Perceiver does not take predictive uncertainty and calibration into account. The Perceiver also generalizes its performance on three datasets, three models, one evaluation metric, and one hyper-parameter setting. Worst of all, the Perceiver's relative performance improvement against other models is marginal. Furthermore, its reduction of architectural prior is not substantial; is not equivalent to its quality. Thereby, I invented five mutations of the Perceiver, the Uncertainty-Aware Perceivers, that obtain uncertainty estimates and measured their performance on three metrics. Experimented with CIFAR-10 and CIFAR-100, the Uncertainty-Aware Perceivers make considerable performance enhancement compared to the Perceiver.
\end{abstract}

\section{Introduction}

The Perceiver (Jaegle et al. [2]) employs a single Transformer-based architecture to manipulate inconsistent arrangements of different modalities. For example, Transformers (Vaswani et al. [14]) comparably make fewer modality-specific assumptions about their inputs' grid structure than 2D convolution operations do. With its Transformer layers, the Perceiver not only maintains the expressivity and flexibility for arbitrary input settings but also handles high-dimensional inputs. 

To be specific, the Perceiver builds an attention bottleneck by utilizing a tiny set of latent units. This bottleneck abolishes the quadratic scaling problem of a self-attention module of a conventional Transformer and removes cohesion between the network depth and the input’s size to build very deep models. The Perceiver transmits its limited capacity to the most relevant inputs by attending to the inputs repetitively.

However, spatial or temporal information is essential for many modules to differentiate input from one modality or another in multi-modal contexts. Thus, the Perceiver links positional and modality-specific features to every input element to reimburse for the absence of explicit structures in its architecture. This association is analogous to tagging input units with a high-fidelity representation of position and modality.

Admittedly, the Perceiver's performance is comparable to ResNet-50 and ViT when trained on ImageNet for multi-variate classification. It also performs competitively on AudioSet's audio and video sound event classification task and ModelNet-40 point cloud classification. However, moving forward would certainly require more evidence and thought.

First, the Perceiver does not consider probability when the model makes unsubstantiated guesses. Hence, the Perceiver does not estimate well-calibrated uncertainty and tends to produce overconfident predictions. Overconfident, incorrect predictions can be destructive or insulting; thus, proper uncertainty quantification has a fatal effect on the Perceiver.

Second, the Perceiver rashly dismisses the fact that its performance on three datasets can not be generalized to all datasets, hyper-parameters, models, and metrics. Surely, the Perceiver has higher accuracy than ViT and ResNet-50 (He et al. [9]) in ImageNet, AudioSet, and ModelNet-40. Nonetheless, does it perform well on a different dataset? Are hyper-parameters of three models optimal? Does it operate better than Coca (Yu et al. [5]), CoAtNet-7 (Dai et al. [4]), ViT-G/14, and other state-of-art models do? Does it have better performance on Bayesian or frequentist performance metrics like Negative Log-Likelihood, Brier Score, etc.?

Third, the Perceiver's performance is not substantial enough to prove its effectiveness. For instance, the Perceiver has 0.1\% higher validation accuracy than ViT-B-16 has. An increase in 0.1\% of validation accuracy can be achieved by mere hyper-parameter or hidden layer size tuning. In addition, CNN-14 has a higher mAP than the Perceiver has on AudioSet; PointNet++ outperforms the Perceiver on ModelNet40. Accordingly, the Perceiver does not excel. 

Lastly, the reduction of architectural prior in the Peceiver is neither substantial nor equivalent to its quality. Specifically, the best-performing Peceiver model on ImageNet uses Fourier feature positional embedding, which is inductive bias. Additionally, removing this positional assumption deteriorates the Perceiver's validation accuracy on permuted ImageNet. Moreover, some inductive biases in the model pale in comparison with its performance, latency, and throughput.

In brief, the Perceiver's strengths are unconvincing due to the lack of pieces of evidence mentioned above. To have a nuanced improvement, the Perceiver needs to consider other factors that can lead to better performance and architecture. By examining all the various angles and factors involved with the Perceiver, it can be concluded that the Perceiver can be improved. 

Therefore, I added Negative Log-Likelihood and Expected Calibration Error as evaluation criteria and performed intensive hyper-parameter tuning. I also looked for new ideas from everywhere and innovated the Uncertainty-Aware Perceivers to mitigate the Perceiver drawbacks. The Uncertainty-Aware Perceivers estimate predictive uncertainty in the training or test stage to reduce uncertainty's impact on optimization. The Uncertainty-Aware Perceivers have five variants: Deep-Perceiver, SWA-Perceiver, Snap-Perceiver, Fast-Perceiver, and MC-Perceiver.

Trained on CIFAR-10 and CIFAR-100, the Uncertainty-Aware Perceivers achieve better performance than the Perceivers, except for the MC-perceiver on CIFAR-10; the Deep Perceiver even outperforms ViT and ResNet-50 on the two datasets.

\section{Related Work}
\subsection{Perceiver}

\subsubsection{Architecture}
The components of the architecture of the Perceiver are two-fold: a cross-attention module and a Transformer tower. The inputs of these two ingredients are byte array and latent array. The byte array is decided by the input data, while the latent array is a hyperparameter. The cross-attention module takes the byte array and latent array to build the latent array. On the other hand, the Transformer tower outputs the latent array with inputs of the latent array. The Perceiver leverages the cross-attention and the Transformer in alternation. The Perceiver also optionally shares weights among each occurrence of the Transformer tower (Fig. 1).

\begin{figure}
    \centering
    \includegraphics[width=.8\textwidth]{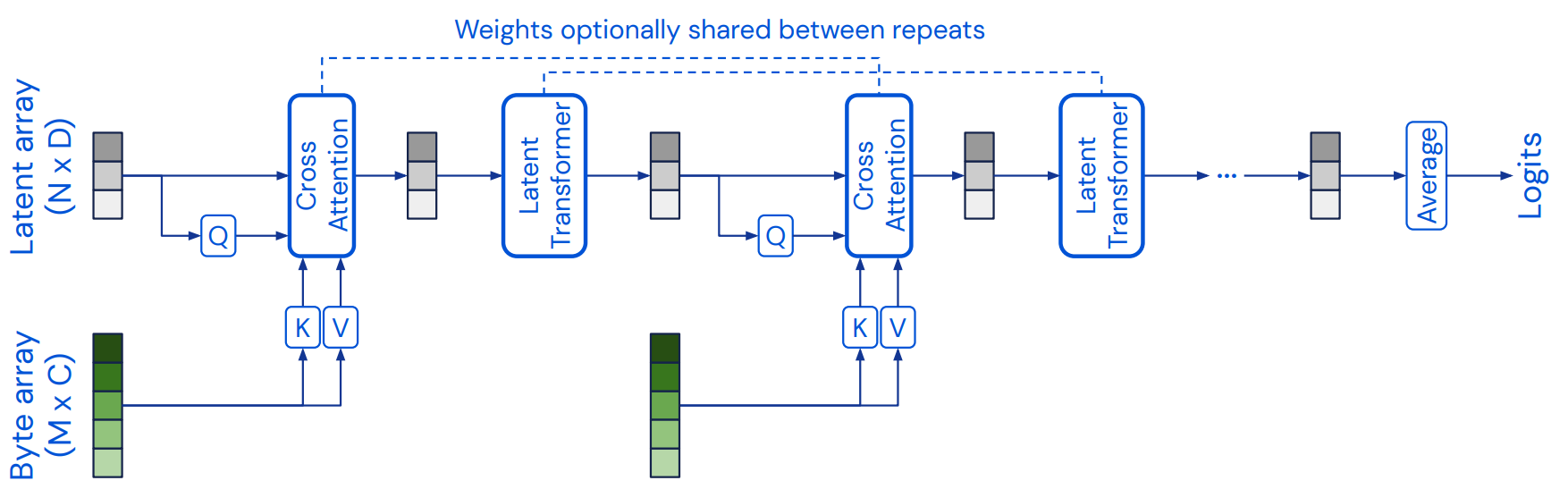}
    \caption{The Perceiver recursively attends to the input byte array by switching between cross-attention and latent self-attention blocks. }
\end{figure}
\subsubsection{Cross-Attention}
Cross-attention utilizes query-key-value (QKV) attention. However, the sequence length (M) of the input is typically very large. Thereby, the cross-attention applies attention directly to the inputs by adding an asymmetry to the attention. Succinctly, K and V are projections of the input byte array; however, Q forecasts a learned latent array that has index dimension N $<<$ M, where N is a hyperparameter. Therefore, the cross-attention operation has O(MN) complexity.

\subsubsection{Latent Transformer}
The latent Transformer has a complexity of O($N^2$). The low cost of the latent Transformer authorizes the Perceiver to have much deeper layers than a traditional Transformers, which has complexity O(LM$^2$) where N $<<$ M. Additionally, the latent Transformer uses the GPT-2 architecture (Radford et al. [6]), that makes use of the Transformer decoder.

\subsubsection{Iterative Cross-Attention}
The Perceiver consists of multiple cross-attend layers for its latent arrays to extract information repetitively. These redundant layers balance expensive but informative against cheaper but redundant cross-attends. Although more cross-attends increase memory usage, they lead to better performance.

\subsubsection{Weight Sharing}
The Perceiver increases the parameter efficiency of the model by sharing weights between the corresponding blocks of each latent Transformer and between cross-attends. Latent self-attention blocks can still be shared if only a single cross-attend is used. The weight sharing reduces 90\% of the number of parameters of the conventional Transformer. Strangest of all, the weight sharing boosts validation performance while reducing overfitting. The resulting architecture has the functional form of an RNN.

\subsubsection{Positional Encoding}
The Perceiver uses Fourier feature positional embeddings. The Fourier feature positional encodings directly represent the temporal and spatial structure of the input data. The Fourier feature positional encodings control frequency bands independent of the cutoff frequency. The Fourier feature positional encodings uniformly sample all frequencies up to a target resolution.

\section{Methods}
\subsection{Deep-Perceiver}

The Deep-Perceiver uses a randomization-based ensemble since it is better fitted for parallel and distributed computation like the Deep Ensemble (Lakshminarayanan et al. [13]). Deep-Perceiver employs the entire training dataset for training as more data points increase deep neural networks' performance. The Deep-Perceiver employs random shuffling of the data point and random initialization of its parameters. The Deep-Perceiver treats the ensemble as a uniformly weighted mixture model and combines the predictions as ${p(y|x) = M^{-1}\Sigma_{m=1}^{M} p_{\theta_m}(y|x, \theta_m)}$. This procedure is equivalent to averaging the predicted probabilities. Before averaging, the Deep-Perceiver's confidences are calibrated with the Temperature Scaling. (Guo et al. [10]).

\begin{algorithm}
\caption{The Deep-Perceiver's training}\label{alg:cap}
\begin{algorithmic}
\Require{Let each neural network parametrize a distribution over the outputs (${p_{\theta}(y|x)}$ and M = 4)}

\Ensure {Initialize ${\theta_{1}}$, ${\theta_{2}}$,..., ${\theta_{M}}$ randomly}
  \For{\texttt{m =1: M}}
    \State {Sample data point ${n_{m}}$ randomly for each net}
    \State {Minimize ${l(\theta_m, x_{n_m}, y_{n_m})}$ }
  \EndFor
\end{algorithmic}
\end{algorithm}

\subsection{SWA-Perceiver}
The SWA-Perceiver applies a cyclical learning rate and uncomplicated mean of multiple points along the trajectory of AdamW, similar to the Stochastic Weight Average (Izmailov et al. [11]). ${\hat{w}}$ -- the SWA-Perceiver Model -- can be trained with the conventional training procedure for a reduced or full training budget. The SWA-Perceiver can stop the training early without changing the learning rate schedule. Starting from ${\hat{w}}$, the SWA-Perceiver starts training using a cyclical learning rate to capture the models ${w_i}$ that is the minimum value of
the learning rate. As a result, the SWA-Perceiver leads to better generalization than the Perceiver. It also guides to wider and flatter than the optima found by AdamW.
\begin{algorithm}
\caption{The SWA-Perceiver's training}\label{alg:cap}
\begin{algorithmic}
\Require{weights ${\hat{w}}$, LR bounds ${\alpha_1, \alpha_2}$,}
\State{cycle length ${c}$ (for constant learning rate ${c}$ = 1),}
\State{number of iterations ${n}$}

\Ensure {${w_{SWA}}$}
  \State{${w \gets \hat{w}}$ {Initialize weights with ${\hat{w}}$}}
  \State{${w_{SWA} \gets w}$}
  \For{\texttt{${i \gets}$ 1,2,...,n }}
    \State {${a \gets a(i)}$ \Comment{Calculate LR for the iteration}}
    \State {${w \gets w - a L_i (w)}$ \Comment{Stochastic gradient update}}
    \If{$mod(i, c) = 0$}
        \State {${n_{models} \gets {i}/{c}}$ \Comment{Number of models}}
        \State {${w_{SWA} \gets \frac{w_{SWA} * n_{models} + w}{n_{models}+1}}$ \Comment{Update average}}
    \EndIf
  \EndFor
\end{algorithmic}
\end{algorithm}
\subsection{Snap-Perceiver}

\begin{figure}
    \centering
    \includegraphics[width=.8\textwidth]{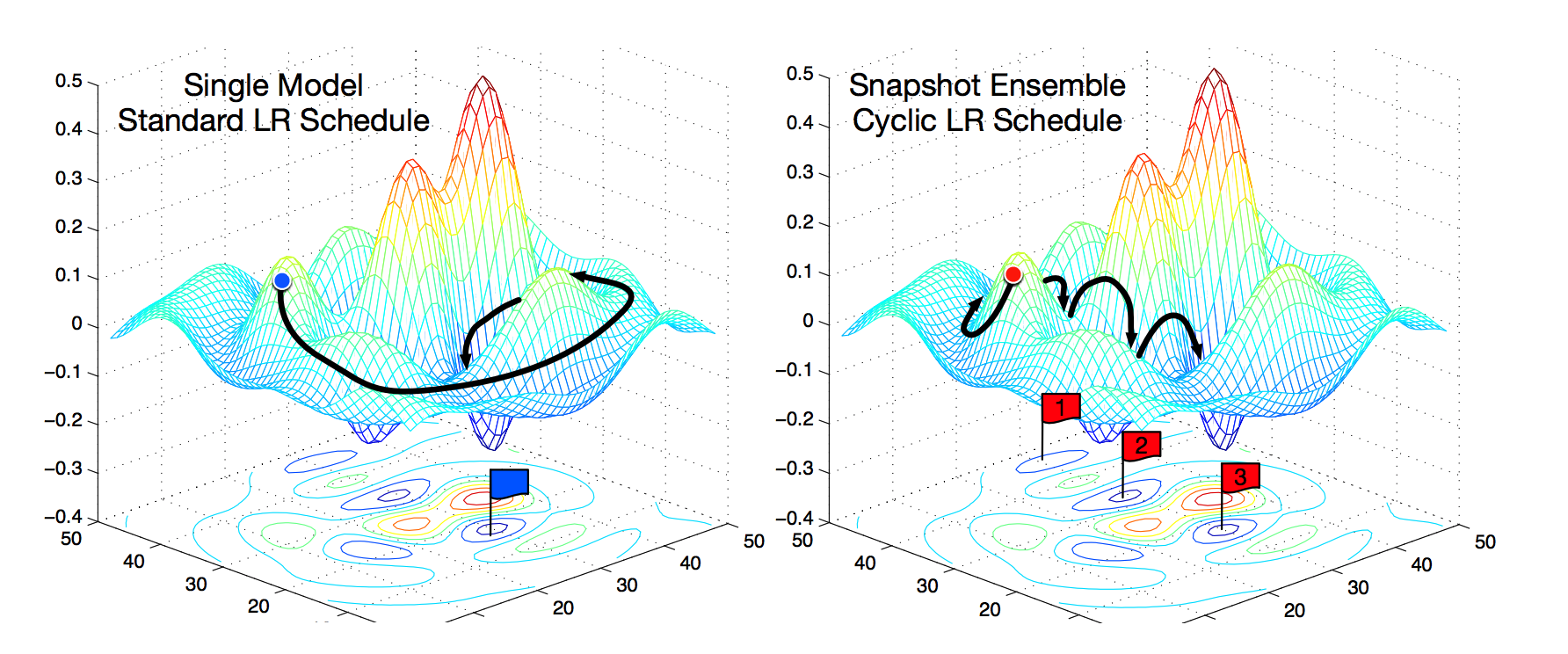}
    \caption{\textbf{Left}: SGD optimization using a conventional learning rate schedule. \textbf{Right}: Illustration of Snap-Perceiver using AdamW. }
\label{fig:2}
\end{figure}

The Snap-Perceiver utilizes Snapshot Ensemble (Huang et al. [12]) to achieve the paradoxical goal of ensembling the Perceiver without additional training cost. The Snap-Perceiver trains a single neural network, converging to various local minima along its optimization path. For fast convergence, the Snap-Perceiver leverages cyclic learning rate schedules. At the end of each training cycle, the Snap-Perceiver gets to a local minimum concerning the training loss. Thus, the Snap-Perceiver takes a "snapshot" of its weights before increasing the learning rate. At its test time, the Snap-Perceiver averages the last m model’s Softmax output (Fig. \ref{fig:2}).

\subsection{Fast-Perceiver}

The Fast-Perceiver casts Fast Ensemble (Garipov et al. [7]) to find high-accuracy pathways among modes. The Fast Ensemble finds simple curves to connect optima of loss functions. Inspired by this geometric insight, the Fast-Perceiver trains ensembles in the time required to train a single model. For example, the Fast-Perceiver initializes a copy of the network with weights w set equal to the weights of the trained network $\hat{w}$. Then, the Fast-Perceiver adapts a cyclical learning rate schedule ${a(\cdot)}$ to force w to move away from $\hat{w}$ without performance degradation.

\begin{figure}
    \centering
    \includegraphics[width=.8\textwidth]{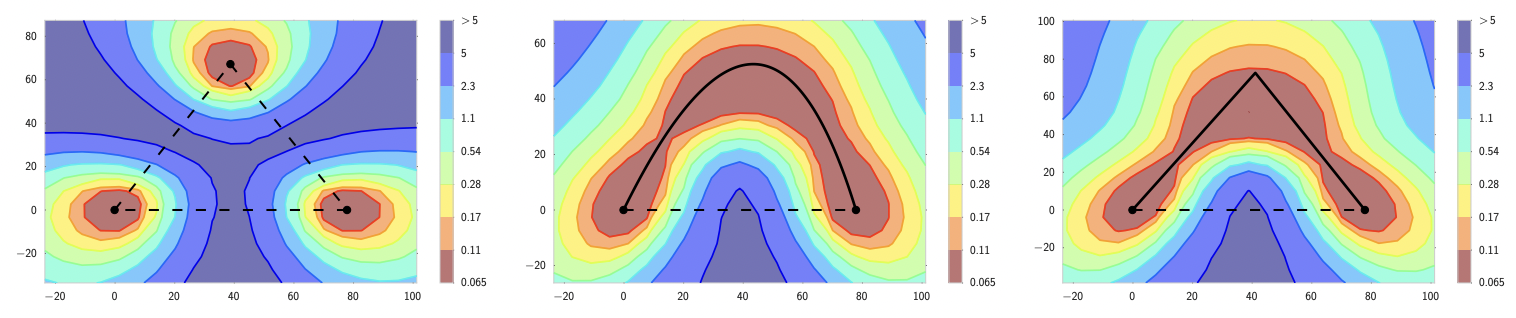}
    \caption{\textbf{Left}: Optima of three distinctly trained networks. \textbf{Middle} and \textbf{Right}: A quadratic Bezier curve used by the Fast-Perceiver, connecting the lower two optima. }
\end{figure}
\subsection{MC-Perceiver}

The MC-Perceiver leverages Monte Carlo Dropout (Gal et al. [3]) training to approximate Bayesian inference in the Perceiver. The MC-Perceiver withdraws abandoned information from the Perceiver to model uncertainty. The MC-Perceiver randomly set pixels of input images to 0 with a certain probability ($\delta$) on its training and test time. In its test stage, the MC-Perceiver $n$ makes dropped-out samples and averages them to use as a prediction. This process alleviates the difficulty of illustrating uncertainty in the Perceiver (Fig. 4).

\begin{figure}
    \centering
    \includegraphics[width=.8\textwidth]{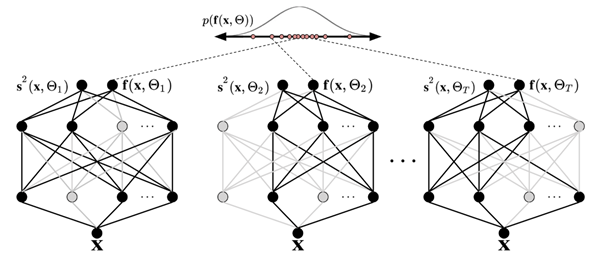}
    \caption{ Illustration of MC Dropout. }
\end{figure}

\section{Experiments}
\begin{table}
  \caption{Reproduction of the Perceiver's paper on CIFAR-10 with Accuracy (the higher, the better) and Negative Log-Likelihood (NLL) (the lower, the better). (2D) implies the usage of 2D convolutions exploits domain-specific grid structure. (FF) is an acronym for Fourier Feature positional embedding, while (L) means Learnable positional embedding. }
  \label{sample-table}
  \centering
  \begin{tabular}{lll}
    \toprule
    \cmidrule(r){1-2}
    Model     & Accuracy (\%)     & NLL \\
    \midrule
    Perceiver (2D) & 0.975  & 0.081     \\
    Perceiver (FF) &  0.982 & 0.067     \\
    Perceiver (L) & 0.954  & 0.139     \\
    ViT & 0.973  &  0.083    \\
    ResNet-50 &  0.9463 &  0.266    \\

    \bottomrule
  \end{tabular}
\end{table}
\begin{table}
  \caption{Reproduction of the Perceiver's paper on CIFAR-100. }
  \label{sample-table2}
  \centering
  \begin{tabular}{llll}
    \toprule
    \cmidrule(r){1-2}
    Model     & Accuracy (\%)     & NLL \\
    \midrule
    Perceiver (2D) & 0.849  & 0.545     \\
    Perceiver (FF) & 0.856 & 0.493     \\
    Perceiver (L) & 0.787  & 0.729     \\
    ViT & 0.84  &  0.864    \\
    ResNet-50 &  0.9 &  0.39    \\

    \bottomrule
  \end{tabular}
\end{table}
\begin{table}
  \caption{ Uncertainty-Aware Perceivers on CIFAR-10 with Expected Calibration Error (ECE, the lower the better).}
  \label{sample-table2}
  \centering
  \begin{tabular}{llll}
    \toprule
    \cmidrule(r){1-2}
    Model     & Accuracy (\%)     & NLL & ECE \\
    \midrule
    Perceiver (2D) & 0.975  & 0.081 & 0.032   \\
    Perceiver (FF) &  0.982 & 0.067  & 0.02  \\
    Perceiver (L) & 0.954  & 0.139  & 0.056  \\
    Deep-Perceiver & 0.99  & 0.03 &  0.015  \\
    SWA-Perceiver & 0.982  & 0.061 &  0.02  \\
    Snap-Perceiver & 0.985  & 0.051 & 0.018   \\
    Fast-Perceiver & 0.985  & 0.152 & 0.015   \\
    MC-Perceiver & 0.975  & 0.081 &  0.031  \\
    \bottomrule
  \end{tabular}
\end{table}
\begin{table}
  \caption{ Uncertainty-Aware Perceivers on CIFAR-100.}
  \label{sample-table2}
  \centering
  \begin{tabular}{llll}
    \toprule
    \cmidrule(r){1-2}
    Model     & Accuracy (\%)     & NLL & ECE \\
    \midrule
    Perceiver (2D) & 0.849  & 0.545 & 0.152   \\
    Perceiver (FF) & 0.856 & 0.493 & 0.152   \\
    Perceiver (L) & 0.787  & 0.729 & 0.205   \\
    Deep-Perceiver & 0.913  & 0.293 & 0.102   \\
    SWA-Perceiver & 0.861  & 0.502 & 0.144   \\
    Snap-Perceiver & 0.864  & 0.443 & 0.152   \\
    Fast-Perceiver &  0.878 & 0.965  &  0.116  \\
    MC-Perceiver & 0.859  & 0.46 &  0.153  \\
    \bottomrule
  \end{tabular}
\end{table}

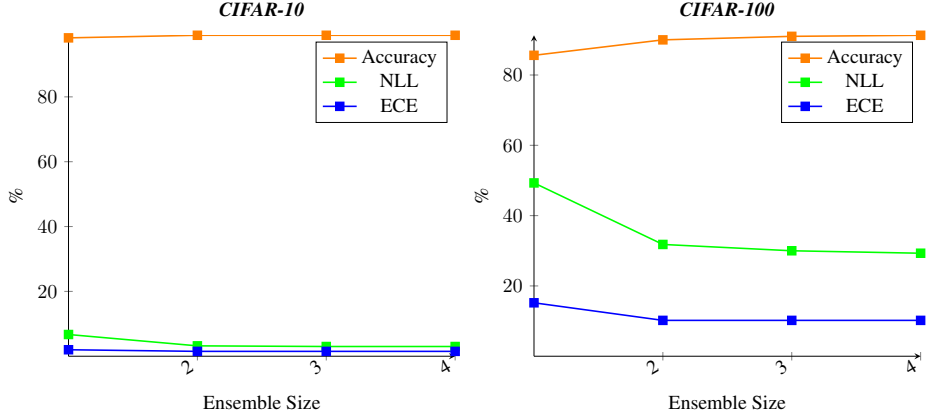
\begin{figure}[h!]
\begin{tikzpicture}[scale=.75]
\begin{axis}[
axis lines=middle,
ymin=0,
    x label style={at={(axis description cs:0.5,-0.1)},anchor=north},
    y label style={at={(axis description cs:-0.1,.5)},rotate=90,anchor=south},
title={\textit{\textbf{CIFAR-10}}},
    xlabel=Ensemble Size,
  ylabel=\%,
  xticklabel style = {rotate=30,anchor=east},
   enlargelimits = false,
  xticklabels from table={cifar10}{Ensemble_Size},xtick=data]
\addplot[orange,thick,mark=square*] table [y=Accuracy,x=X]{cifar10};
\addlegendentry{Accuracy}
\addplot[green,thick,mark=square*] table [y= NLL,x=X]{cifar10};
\addlegendentry{NLL}]
\addplot[blue,thick,mark=square*] table [y= ECE,x=X]{cifar10};
\addlegendentry{ECE}]
\end{axis}
\end{tikzpicture}
\begin{tikzpicture}[scale=.75]
\begin{axis}[
axis lines=middle,
ymin=0,
    x label style={at={(axis description cs:0.5,-0.1)},anchor=north},
    y label style={at={(axis description cs:-0.1,.5)},rotate=90,anchor=south},
title={\textit{\textbf{CIFAR-100}}},
    xlabel=Ensemble Size,
  ylabel=\%,
  xticklabel style = {rotate=30,anchor=east},
   enlargelimits = false,
  xticklabels from table={cifar100}{Ensemble_Size},xtick=data]
\addplot[orange,thick,mark=square*] table [y=Accuracy,x=X]{cifar100};
\addlegendentry{Accuracy}
\addplot[green,thick,mark=square*] table [y= NLL,x=X]{cifar100};
\addlegendentry{NLL}]
\addplot[blue,thick,mark=square*] table [y= ECE,x=X]{cifar100};
\addlegendentry{ECE}]
\end{axis}
\end{tikzpicture}
\caption{The Deep-Perceiver's predictive performance as a function of ensemble size on CIFAR-10 and CIFAR-100.}
\end{figure}

To measure the Perceiver's generalization performance, I reproduced the Perceiver, ViT, and ResNet-50 as baselines on CIFAR-10 (Tab. 1) and CIFAR-100 (Tab. 2) on an RTX 3090 D6X 24GB. Empirically, batch size of 4 and learning rate of 5e-6 had the best performance for the Perceiver. Although the Perceiver with Fourier feature positional embedding had the best performance among the three models on CIFAR-10, ViT conspicuously surpassed the Perceiver and ResNet-50 on CIFAR-100. 

With the same hyper-parameters above, I ran experiments on the Deep-Perceiver, SWA-Perceiver, Snap-Perceiver, Fast-Perceiver, and MC-Perceiver. On CIFAR-10 (Tab. 3), all the models except for the MC-Perceiver transcended the baselines. Although all five variants exceeded the Perceiver and ResNet-50 on CIFAR-100 (Tab. 4), only Deep-Perceiver outperformed ViT on CIFAR-100 (Fig. 3). 

I averaged Softmax outputs of 4 independently trained Perceivers with randomly initialized parameters for the Deep-Perceiver. However, logits before the Softmax outputs of the Perceivers were scaled using temperature, $T$. $T$ was determined by the Negative Log-Likelihood between the actual labels and logits using the Nelder-Mead optimizer. The Deep-Perceiver was run with an ensemble size of 1, 2, 3, and 4. More ensemble yielded better Accuracy, ECE, and NLL (Fig. 4).

For the SWA-Perceiver, 10 optimization steps were used to convert the learning rate of a constant value of 5e-6 to 2e-6. At the end of every 5 optimization steps, snapshots of the weights were added to the SWA running average. In contrast, the Fast-Perceiver used 4 cycles to anneal the learning rate from 5e-6 to 5e-7.

The cyclic learning rate was scheduled for the Snap-Perceiver with the equation, \begin{center}
$a(t) = \frac{a_0}{2} (cos(\frac{\pi mod(t-1, [ T/M ])}{[T/M]}) + 1)$,
\end{center}

where $a_0$ is the initial learning rate. In comparison, the MC-Perceiver used a constant learning rate schedule that generated 30 samples before taking the average for a prediction.

\section{Conclusion}

In summary, the Perceiver's effectiveness and logical coherence have proved unconvincing. To strengthen it, the Perceiver needs to contain more evaluation metrics and probabilistic optimizations. On the other hand, my novel Uncertainty-Aware Perceiver takes calibrated uncertainty into account while retaining high scalability and skyrocketing performance. In particular, the Deep Perceiver outperforms the Perceiver, ViT, and ResNet-50 on both CIFAR-10 and CIFAR-100.

In future work, I would like to pre-train the Uncertainty-Aware Perceiver in a feature-based and fine-tuning approach. Succinctly, I wish to add a denoised Variational AutoEncoder or Generative Adversarial Network as a pre-training stage to further capture uncertainties associated with its input. Moreover, I want to convert the Transformer architecture in the Perceiver to Bayesian to minimize its uncertainty. Furthermore, I hope to further judge uncertainty estimates by adding a Brier Score, defined as ${BS = K^{-1}\Sigma_{K=1}^{K}(t_k^{*}-p(y=k|x^{*}))^{2})}$, where ${t_k^{*} = 1}$ if ${k = y^{*}}$ and 0 otherwise.
\pagebreak 
\section*{References}
\label{reference}

{
\small

[1] Hu, P. and Ramanan, D. Bottom-up and top-down reasoning
with hierarchical rectified gaussians. {\it In Proceedings of
IEEE Conference on Computer Vision and Pattern Recognition (CVPR). Advances in Neural Information Processing Systems 7}. pp.\ 609--616. Cambridge, MA: MIT Press.

[2] Jaegle, A., Felix G., Brock A., Zisserman A., Vinyals O. \& Carreira J. (2021) Perceiver: General Perception with Iterative Attention. {\it In Proceedings of International Conference on Machine Learning}, PMLR 139. London, UK.

[3] Gal, Y. (2016) Dropout as a Bayesian Approximation: Representing Model Uncertainty in Deep Learning. {\it Proceedings of the 33 rd International Conference on Machine Learning}, JMLR: W\&CP volume 48. NY, USA.

[4] Dai, Z., Liu, H., Le, Q., \& Tan, M. (2021) CoAtNet: Marrying Convolution and Attention for All Data Sizes. {\it Preprint}.

[5] Yu, Z. and Wang, Z. (2022) CoCa: Contrastive Captioners are Image-Text
Foundation Models. {\it Preprint}.

[6] Radford, A., Wu, J., Child, R., Luan, D., Amodei, D., \& Sutskever, I. (2019) Language Models are Unsupervised Multitask Learners. {\it OpenAI}, San Francisco, California, United States.

[7] Garipov, T., Izmailov, P., Podoprikhin, D., Vetrov, D., \& Wilson, A. (2018) Loss Surfaces, Mode Connectivity, and Fast Ensembling of DNNs. {\it 32nd Conference on Neural Information Processing Systems}. Montréal, Canada.

[8] Dosovitskiy, A., Beyer, L., Kolesnikov, A., Weissenborn, D., Zhai, X., Unterthiner, T., Dehghani, M., Minderer, M., Heigold, G., Gelly, S., Uszkoreit, J., \& Houlsby, N. (2021) Transformers for Image Recognition at Scale. {\it The International Conference on Learning Representations (ICLR)}. 

[9] He, K., Zhang, X., Ren, R., \& Sun, J. (2015) Deep Residual Learning for Image Recognition. {\it The IEEE / CVF Computer Vision and Pattern Recognition Conference (CVPR)}.

[10] Guo, C., Pleiss, G., Sun, Y., \& Weinberger, K. (2017) On Calibration of Modern Neural Networks. {\it Proceedings of the 34 th International Conference on Machine Learning}. PMLR 70. Sydney, Australia.

[11] Izmailov, P., Podoprikhin, D., Garipov, T., \& Vetrov, D., Wilson, A. (2018) Averaging Weights Leads to Wider Optima and Better Generalization. {\it Conference on Uncertainty in Artificial Intelligence}.

[12] Huang, G., Li, Y., Pleiss, G., Liu, J., \& Weinberger, J. (2017) Snapshot Ensembles: Train 1, Get M for Free. {\it The International Conference on Learning Representations (ICLR)}. 

[13] Lakshminarayanan, B., Pritzel, A., \& Blundell, C. (2017) Simple and Scalable Predictive Uncertainty Estimation using Deep Ensembles. {\it 31st Conference on Neural Information Processing Systems}. Long Beach, CA, USA.

[14] Vaswani, A., Shazeer, N., Parmar, N., Uszkoreit, J., Jones, L., Gomez, A., Kaiser, L., \& Polosukhin, I. (2017) Attention Is All You Need. {\it 31st Conference on Neural Information Processing Systems}. Long Beach, CA, USA.

\end{document}